\documentclass[lettersize,journal]{IEEEtran}
\usepackage{amsmath,amsfonts}
\usepackage{algorithmic}
\usepackage{algorithm}
\usepackage{array}
\usepackage[caption=false,font=normalsize,labelfont=sf,textfont=sf]{subfig}
\usepackage{textcomp}
\usepackage{stfloats}
\usepackage{url}
\usepackage{verbatim}
\usepackage{graphicx}
\usepackage{cite}
\usepackage{makecell}
\usepackage{xcolor}
\usepackage{colortbl}
\usepackage{amssymb}
\usepackage{booktabs}
\usepackage{bbding}
\usepackage{tabularx}
\usepackage{amsmath}
\usepackage{multirow}
\usepackage{enumitem}
\usepackage{tcolorbox}
\usepackage[colorlinks=true,
            linkcolor=blue,       
            anchorcolor=red,     
            citecolor=blue,       
            urlcolor=black,
            ]{hyperref}
\begin{document}

\title{From Ideal to Real: Unified and Data-Efficient \\ Dense Prediction for Real-World Scenarios}


\author{Changliang Xia$^*$,
        Chengyou Jia$^*$,
        Zhuohang Dang,
        Minnan Luo$^\dagger$,\\
        Zhihui Li,
        and Xiaojun Chang
        \thanks{$^*$Equal Contribution.}
        \thanks{$^\dagger$Corresponding author: Minnan Luo.}
        \thanks{ Changliang Xia, Chengyou Jia, Zhuohang Dang and Minnan Luo, are with the School of Computer Science and Technology,
        Xi’an Jiaotong University, Xi’an 710049, China (e-mail: 202066@stu.xjtu.edu.cn; cp3jia@stu.xjtu.edu.cn; dangzhuohang@
        stu.xjtu.edu.cn; minnluo@xjtu.edu.cn;).}
        \thanks{Zhihui Li and Xiaojun Chang are with the School of Information Science and Technology, University of Science and Technology of China, Hefei 230026, China (e-mail: lizhihuics@ustc.edu.cn; xiaojun.chang@uts.edu.au.)}.
        \thanks{This work has been submitted to the IEEE for possible publication. Copyright may be transferred without notice, after which this version may no longer be accessible.}
}



\maketitle

\begin{abstract}
Dense prediction tasks hold significant importance of computer vision, aiming to learn pixel-wise annotated labels for input images. Despite advances in this field, existing methods primarily focus on idealized conditions, exhibiting limited real-world generalization and struggling with the acute scarcity of real-world data in practical scenarios. 
To systematically study this problem, we first introduce \textbf{DenseWorld}, a benchmark spanning a broad set of 25 dense prediction tasks that correspond to urgent real-world applications, featuring unified evaluation across tasks.
We then propose \textbf{DenseDiT}, which exploits generative models' visual priors to perform diverse real-world dense prediction tasks through a unified strategy.
DenseDiT combines a parameter-reuse mechanism and two lightweight branches that adaptively integrate multi-scale context.
This design enables DenseDiT to achieve efficient tuning with less than 0.1\% additional parameters, activating the visual priors while effectively adapting to diverse real-world dense prediction tasks.
Evaluations on DenseWorld reveal significant performance drops in existing general and specialized baselines, highlighting their limited real-world generalization. In contrast, DenseDiT achieves superior results using less than 0.01\% training data of baselines, underscoring its practical value for real-world deployment.
\end{abstract}


\section{Introduction}
\IEEEPARstart{D}ense prediction tasks~\cite{zhou2024rafpn, lu2024prompt}, such as segmentation and depth estimation, represent a class of fundamental computer vision problems that aim to learn mappings from input images to pixel-wise annotated labels. These tasks are critical for numerous applications, such as medical imaging~\cite{shu2022cross}, autonomous driving~\cite{caesar2020nuscenes}, remote sensing~\cite{zhao2024rs}, and others. 
While recent advances~\cite{dong2024unidense,kim2024chameleon,kim2023universal} have achieved strong performance through carefully designed strategies, they are mostly developed under idealized conditions~\cite{silberman2012indoor, geiger2013vision} featuring uniform lighting, minimal occlusion and easily accessible, relatively abundant data, resulting in limited generalization to practical dense prediction tasks~\cite{jin2022fives, benz2024omnicrack30k}.

\begin{figure*}[h]
  \centering
 \includegraphics[width=0.85\textwidth]{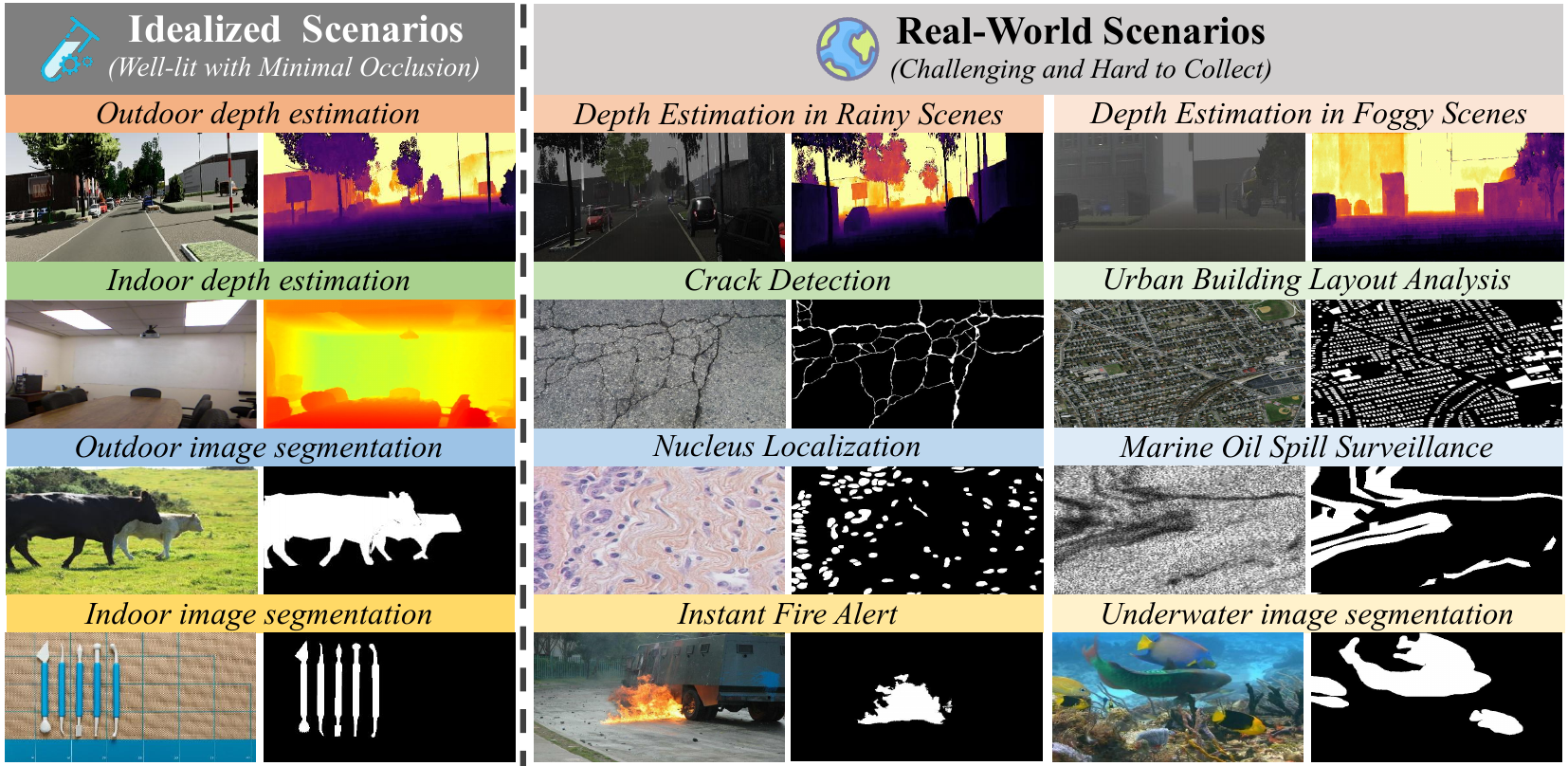}
    \caption{Comparison of idealized vs. real-world dense prediction. (Left) Idealized tasks under controlled conditions with uniform lighting, minimal occlusion, and abundant data. (Right) Real-world tasks exhibiting complex scenes, adverse conditions, and inherent data scarcity, presenting substantially greater challenges.}
  \label{fig:scene_comparison}
\end{figure*}

In contrast to idealized conditions, dense prediction in real-world scenarios offers more practical value. 
These scenarios cover broad applications such as depth estimation in adverse weather to enhance autonomous driving safety, crack detection to enable proactive maintenance, and underwater image segmentation to advance deep-sea exploration.
As illustrated in Fig.~\ref{fig:scene_comparison}, real-world scenarios differ from idealized conditions in two aspects: inherent complexity and data scarcity.
This motivates our core question: \textbf{Can we develop a data-efficient dense prediction model that generalizes effectively across diverse real-world scenarios?}

To this end, we first introduce \textbf{DenseWorld}, a benchmark to advance dense prediction in real-world scenarios. 
The core principle of DenseWorld is to cover a broad spectrum of real-world tasks that are both practical and challenging, while being difficult to collect and annotate at scale.
We carefully curate 25 diverse dense prediction tasks, each corresponding to a distinct real-world scenario. These tasks span a wide range of applications, including ecology~\cite{haug15}, healthcare~\cite{jin2022fives}, infrastructure~\cite{mnih2013machine}, public safety~\cite{chino2015bowfire}, and industrial operations~\cite{qin2022highly}. 
Unlike prior domain-specific benchmarks, DenseWorld establishes a unified evaluation protocol, enabling fair and holistic comparison across heterogeneous real-world tasks. Moreover, it intentionally reflects the inherent data scarcity of real-world applications, where few-shot dense prediction is the norm rather than the exception. In this sense, DenseWorld provides a unified and realistic platform to evaluate dense prediction methods under real-world complexity.

We further propose \textbf{DenseDiT}, a dense prediction framework built upon generative models.
While prior works~\cite{ke2024repurposing,saxena2023monocular} concatenate query and noisy tokens along the channel dimension, they inevitably modify the base model architecture, which compromises its pretrained visual priors.
Moreover, such rigid concatenation restricts flexible interactions between query and noisy tokens.
Instead, DenseDiT adopts a parameter-reuse mechanism to process query tokens and noisy tokens independently within the intact latent space where the visual priors reside.
This independence enables more expressive and flexible interactions via the Multi-Modal Attention (MMA) module~\cite{pan2020multi} of the generative model.
To further enhance generalization across diverse real-world scenarios, we introduce two lightweight branches: a prompt branch, which reuses the text encoder to inject task-specific semantic cues, and a demonstration branch, which reuses the latent encoder to align visual priors with complex scene distributions.
Unlike approaches relying on full training~\cite{ke2024repurposing}, which are fundamentally prone to overfitting and hence impaired generalization given the inherent data scarcity of real-world scenarios, DenseDiT achieves robust performance and generalizes across diverse real-world dense prediction tasks with minimal adaptation.

We conduct comprehensive experiments on the proposed DenseWorld benchmark and common benchmarks to evaluate various approaches.
Extensive quantitative and qualitative results demonstrate that DenseDiT significantly outperforms both general-purpose and task-specific models across all tasks, showcasing strong capabilities for real-world dense prediction.
Compared to existing generative-based dense prediction methods~\cite{ke2024repurposing, le2024maskdiff}, DenseDiT achieves strong performance in more challenging real-world scenarios under limited supervision, highlighting its practical alignment with the data-scarce nature of many real-world applications.
Moreover, the effectiveness of DenseDiT's branches reveals the strong flexible contextual modeling capacity of recent generative models, highlighting promising directions for future research.
In summary, our contributions are as follows:
\begin{itemize}
    \item We introduce DenseWorld, a benchmark for unified evaluation of dense prediction across real-world scenarios. It covers 25 practical applications and faithfully reflects the inherent data scarcity common in real-world settings.
    \item We propose DenseDiT, a unified framework that leverages a parameter-reuse architecture and branch-enhanced adaptation for data-efficient dense prediction across diverse real-world scenarios.
    \item Extensive experiments on the proposed DenseWorld benchmark and common benchmarks demonstrate the superiority of DenseDiT over general-purpose and task-specific baselines, revealing their limited generalization to real-world scenarios and highlighting DenseDiT’s potential for dense prediction in the wild.
\end{itemize}

\section{Related Work}
\subsection{Dense Prediction}
Dense prediction tasks~\cite{zamir2018taskonomy, xia2024vit} such as segmentation and depth estimation have made substantial progress through carefully designed architectures and training pipelines~\cite{dong2024unidense, kim2024chameleon, shi2024unified}. 
However, most existing approaches are developed under idealized conditions, including bright and minimally occluded indoor scenarios~\cite{silberman2012indoor} or outdoor scenarios with favorable weather~\cite{geiger2013vision}, which are typically accompanied by easily obtainable and relatively abundant training data, as shown in the left part of Fig.~\ref{fig:scene_comparison}.
Recent few-shot dense prediction methods~\cite{kim2023universal, hossain2024visual} improve data efficiency via cross-image matching or prompt-based adaptation, but require extensive base-class training and are typically evaluated on controlled benchmark settings. This limits their practicality in diverse, complex, and data-scarce real-world scenarios. To mitigate these limitations, task-specific models~\cite{tao2023convolutional, gasperini2023robust} have been proposed, but they lack generalizability across diverse dense prediction tasks. 
Corresponding datasets are also tailored to specific domains, such as OmniCrack30K~\cite{benz2024omnicrack30k} for crack segmentation and the Global-Scale~\cite{yin2024towards} for road extraction. In this work, we tackle these limitations with a unified and data-efficient framework tailored for real-world dense prediction.
\begin{figure*}[t]
  \centering
  \includegraphics[width=0.85\textwidth]{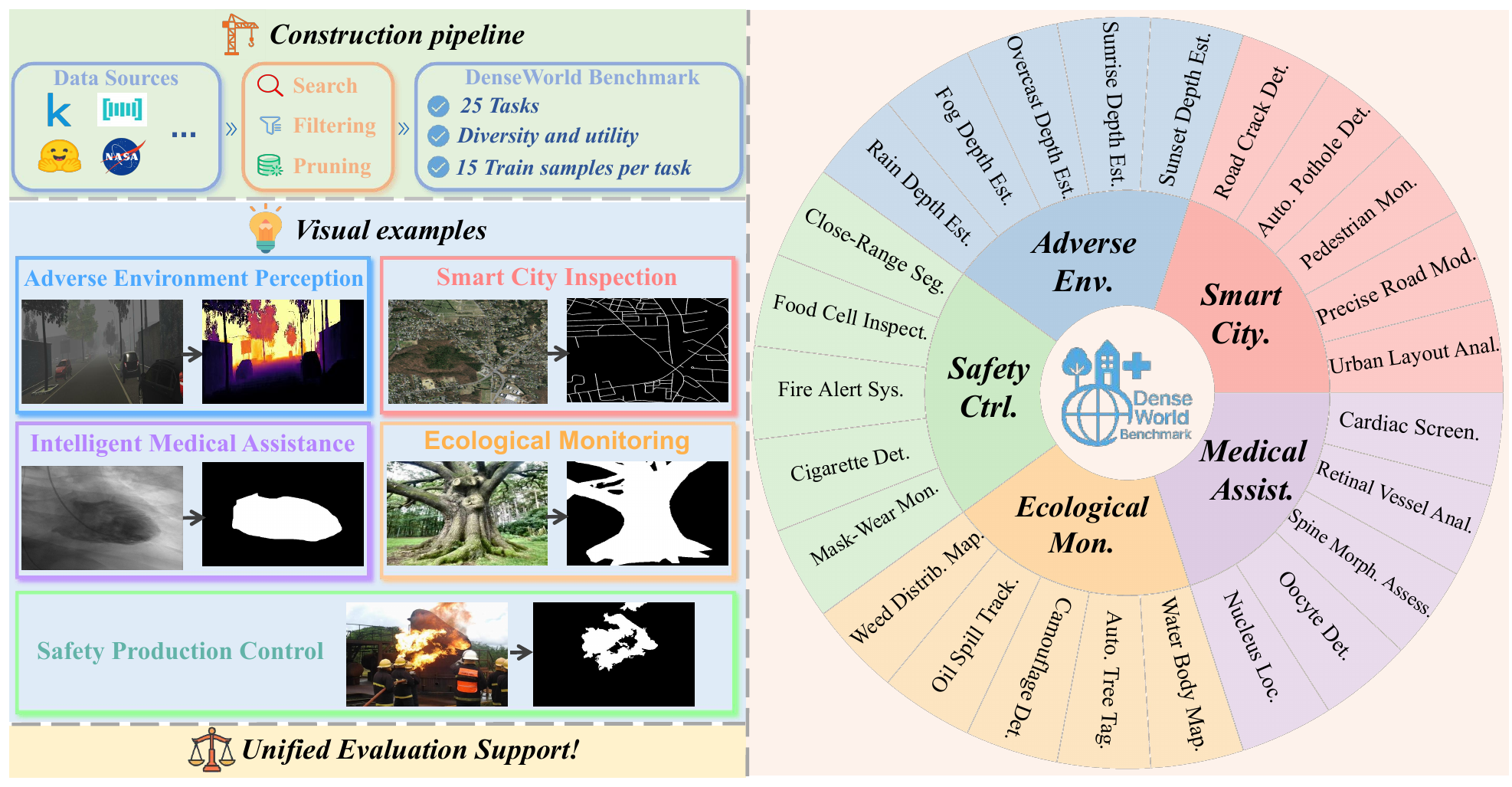}
    \caption{Overview of the DenseWorld benchmark. Upper left: the construction pipeline. Center left: examples of representative tasks across five real-world categories. Lower left: unified evaluation. Right: full taxonomy of 25 dense prediction tasks, each aligned with a practical application scenario.}
  \label{fig:denseworld_intro}
\end{figure*}
\subsection{Diffusion Models For Dense Prediction}
Pretrained generative models have been shown to possess rich visual priors~\cite{bhattad2023stylegan}. These priors have been leveraged for tasks like representation learning~\cite{donahue2019large}. By redefining dense prediction as an image-to-image task, recent work~\cite{ke2024repurposing, zhao2025diception} has explored diffusion models~\cite{rombach2022high, lee2024exploiting} for dense prediction, such as depth estimation and segmentation. In depth estimation, DepthGen~\cite{saxena2023monocular} performs monocular depth estimation using diffusion models, addressing noisy depth data through step-unrolled denoising and infilling strategies. Marigold~\cite{ke2024repurposing} fine-tunes a diffusion model on clean synthetic data to elegantly exploit its visual priors for depth estimation. In segmentation, RefLDM-Seg~\cite{wang2025explore} formulates in-context segmentation as latent mask generation. However, these methods typically require modifying the backbone architecture, which compromises the visual priors. Moreover, they are mostly trained under idealized conditions, lacking generalization to real-world scenarios. In contrast, our work enables diffusion models to generalize across diverse real-world dense prediction tasks with minimal adaptation effort.

\section{DenseWorld Benchmark}
We introduce the \textbf{DenseWorld} benchmark, comprising diverse and practically valuable real-world dense prediction tasks that suffer from limited data, as shown in Fig.~\ref{fig:denseworld_intro}.
\subsection{Benchmark Construction}
An overview of the benchmark construction pipeline is shown in the upper left of Fig.~\ref{fig:denseworld_intro}, comprising three stages: data collection, task filtering, and data pruning. To build DenseWorld, we collect data from various sources, including open platforms (e.g., Kaggle~\cite{kaggle_platform}), academic repositories (e.g., Papers With Code~\cite{pwcode_platform}), and domain-specific portals (e.g., NASA EarthData~\cite{nasa_earthdata_platform}).
Unlike previous benchmarks that emphasize scale, we prioritize the diversity and utility of real-world scenarios. Notably, the benchmark intentionally reflects the inherent data scarcity common in practical applications, such as medical imaging where expert annotations are costly, or industrial monitoring where data acquisition is inherently difficult. This makes the few-shot setting not an artificial constraint, but a realistic simulation of deployment conditions.
Built upon a cleaned collection of source data, DenseWorld covers 25 dense prediction tasks across five categories, each aligned with a critical real-world application.
\subsection{Unified Evaluation Metric}
\label{sec:unified_eval} 
Real-world dense prediction includes diverse tasks such as depth estimation in autonomous driving and crack segmentation in infrastructure inspection, each traditionally evaluated using domain-specific metrics. This heterogeneity makes cross-domain comparison and overall performance assessment challenging. To address this, DenseWorld introduces a unified evaluation framework centered around the \textbf{D/S-Score}, a composite metric that facilitates fair and consistent comparison across tasks.
For regression tasks (e.g., depth estimation), the \textbf{D-Score} integrates five standard metrics (\textit{AbsRel}, \textit{RMSE}, \textit{SqRel}, \textit{RMSE-log}, \textit{Threshold Accuracy}) following established protocols~\cite{eigen2014depth}. For classification tasks (e.g., segmentation), the \textbf{S-Score} combines IoU, PA, and Dice~\cite{cordts2016cityscapes}.
Normalization is essential for meaningful aggregation across these heterogeneous metrics. They differ significantly in scale and optimal direction. Without normalization, metrics with larger numerical ranges would dominate the aggregate, while opposing optimization directions would render cross-task comparison invalid. We therefore apply min-max normalization to project all metrics to a common [0,1] range with a consistent higher-is-better convention:
\begin{equation}
\text{Normalized}(m) =
\begin{cases}
\dfrac{m - m_{\min}}{m_{\max} - m_{\min}}, & \text{for } m \uparrow \ \\
\dfrac{m_{\max} - m}{m_{\max} - m_{\min}}, & \text{for } m \downarrow
\end{cases}
\end{equation}
where $m \uparrow$ denotes that a higher value of metric $m$ is better, and $m \downarrow$ denotes that a lower value is better.

The final D-Score and S-Score are computed as the arithmetic mean of their respective normalized metrics. 
This approach provides a standardized measure for evaluating model performance across diverse real-world dense prediction tasks.
\subsection{Benchmark Analysis}
DenseWorld offers notable advantages over existing benchmarks in real-world diversity and utility, as shown in Table~\ref{tab:benchmark_comparison}. Existing datasets such as Taskonomy~\cite{zamir2018taskonomy}, COCO~\cite{lin2014microsoft}, NYUv2~\cite{silberman2012indoor}, KITTI~\cite{geiger2013vision}, OmniCrack30k~\cite{benz2024omnicrack30k}, and Global-Scale~\cite{yin2024towards} are limited in real-world task coverage or collected under ideal conditions. In contrast, DenseWorld covers 25 practical tasks, each aligned with a real-world application. Moreover, the benchmark intentionally reflects the inherent data scarcity of practical scenarios, where annotated data is naturally limited. The taxonomy is on the right of Fig.~\ref{fig:denseworld_intro}, with more details (e.g., test set size) in the supplementary material.

\begin{table}[htbp]
\caption{Comparison of dense prediction benchmarks.\label{tab:benchmark_comparison}}
\centering
\setlength{\tabcolsep}{2.2pt}
\fontsize{8.2pt}{9pt}\selectfont
\begin{tabular}{lccccc}
\toprule
\textbf{Dataset} & \makecell{Practical\\Tasks} & \makecell{Training\\Samples} & \makecell{Data-\\Efficiency?} & \makecell{Real-World\\Utility?} & \makecell{Unified\\Eval?} \\
\midrule
Taskonomy     & 1  & 4M &\XSolidBrush &  \XSolidBrush & \CheckmarkBold \\
COCO            & 2  & 118K &\XSolidBrush &  \XSolidBrush & \XSolidBrush \\
NYUv2         & 1  & 795 &\XSolidBrush & \XSolidBrush & \XSolidBrush \\
KITTI            & 2  & 86K &\XSolidBrush & \CheckmarkBold & \XSolidBrush \\
Omnicrack30k & 1  & 22K &\XSolidBrush & \CheckmarkBold & \XSolidBrush \\
Global-Scale       & 1  & 2K &\XSolidBrush & \CheckmarkBold & \XSolidBrush \\
\rowcolor{gray!10}
DenseWorld                                & 25 & 15 &\CheckmarkBold & \CheckmarkBold & \CheckmarkBold \\
\bottomrule
\end{tabular}
\end{table}

\section{Method}
Our goal is to reformulate generative models into data-efficient dense predictors capable of generalizing across diverse real-world scenarios.
Section~\ref{sec:preliminary} introduces the generative model with strong visual priors and our model overview.
Section~\ref{sec:Standard_Task} formalizes the dense prediction tasks.
Finally, Section~\ref{sec:DenseDiT} details our proposed method, DenseDiT.
\subsection{Preliminary and Model Overview}
\label{sec:preliminary}
Our approach builds upon the DiT architecture~\cite{peebles2023scalable}, adopted by recent state-of-the-art generative models~\cite{flux2024, esser2403scaling}. 
DiT uses a transformer-based denoising network to refine noisy latent iteratively.
Each block includes a Multi-Modal Attention (MMA) module~\cite{pan2020multi}, enabling effective interaction between noisy latent and conditioning inputs. 

DiT learns a velocity field~\cite{karras2022elucidating} to map noisy samples to clean ones, optimized by minimizing the discrepancy between predicted and ground-truth velocities:
\begin{equation}
\label{eq:velocity_loss}
\text{Loss} = \mathbb{E}_{z_0, t, c} \left\| v_\theta(z_t, t, c) - u(z_t) \right\|_2^2
\end{equation}
where $c$ is the conditioning information, $z_0$ is the clean latent, and $z_t$ is the noisy latent at timestep $t$.

Fig.~\ref{fig:densedit_arch} illustrates our DenseDiT, a unified and data efficient framework for dense prediction. The unification in our approach reflects three key characteristics: unified evaluation across diverse tasks through D/S-Score (Section~\ref{sec:unified_eval}), unified architecture that handles various real-world dense predictions without task-specific modules, and unified training that applies the same objective and optimization to all tasks, contrasting with prior methods requiring customized designs per task~\cite{xu2024matters}.
DenseDiT takes a query image, a textual prompt, and an optional demonstration pair as inputs. A parameter-reused VAE encoder projects the query images into the latent space. Two lightweight auxiliary branches, the prompt and the demonstration branch, provide semantic and visual contextual cues, activated by a task attribute called DAI. This design is inherently compatible with future advances in efficient generation, such as one-step diffusion~\cite{yin2024one}, and is expected to gain further efficiency as these techniques mature. The inputs are jointly processed through MMA~\cite{pan2020multi} modules for dense prediction. Details are provided in Section~\ref{sec:DenseDiT}.
\subsection{Standardizing Task Representation}
\label{sec:Standard_Task}
Diverse data formats in dense prediction pose challenges to unified processing.
Inspired by~\cite{zhao2025diception}, we standardize task representations into RGB format, as most vision models~\cite{dosovitskiy2020image, rombach2022high}, including DiT, are trained on RGB data, which is critical for leveraging visual priors.
First, we align channel dimensions by duplicating single-channel data to ensure compatibility with RGB-based encoders. 
Then, we normalize pixel values to address scale variations across tasks. Specifically, we apply:

\begin{equation}
\label{eq:reg_normalization}
\mathbf{r}_{\mathrm{norm}} = \left( \frac{\mathbf{r} - \mathbf{r}_{\min}}{\mathbf{r}_{\max} - \mathbf{r}_{\min}} - 0.5 \right) \times 2
\end{equation}
where $\mathbf{r}_{\max}$ and $\mathbf{r}_{\min}$ denote the maximum and minimum.

\subsection{DenseDiT}
\label{sec:DenseDiT}
\begin{figure}[t]
  \centering
  \includegraphics[width=\linewidth]{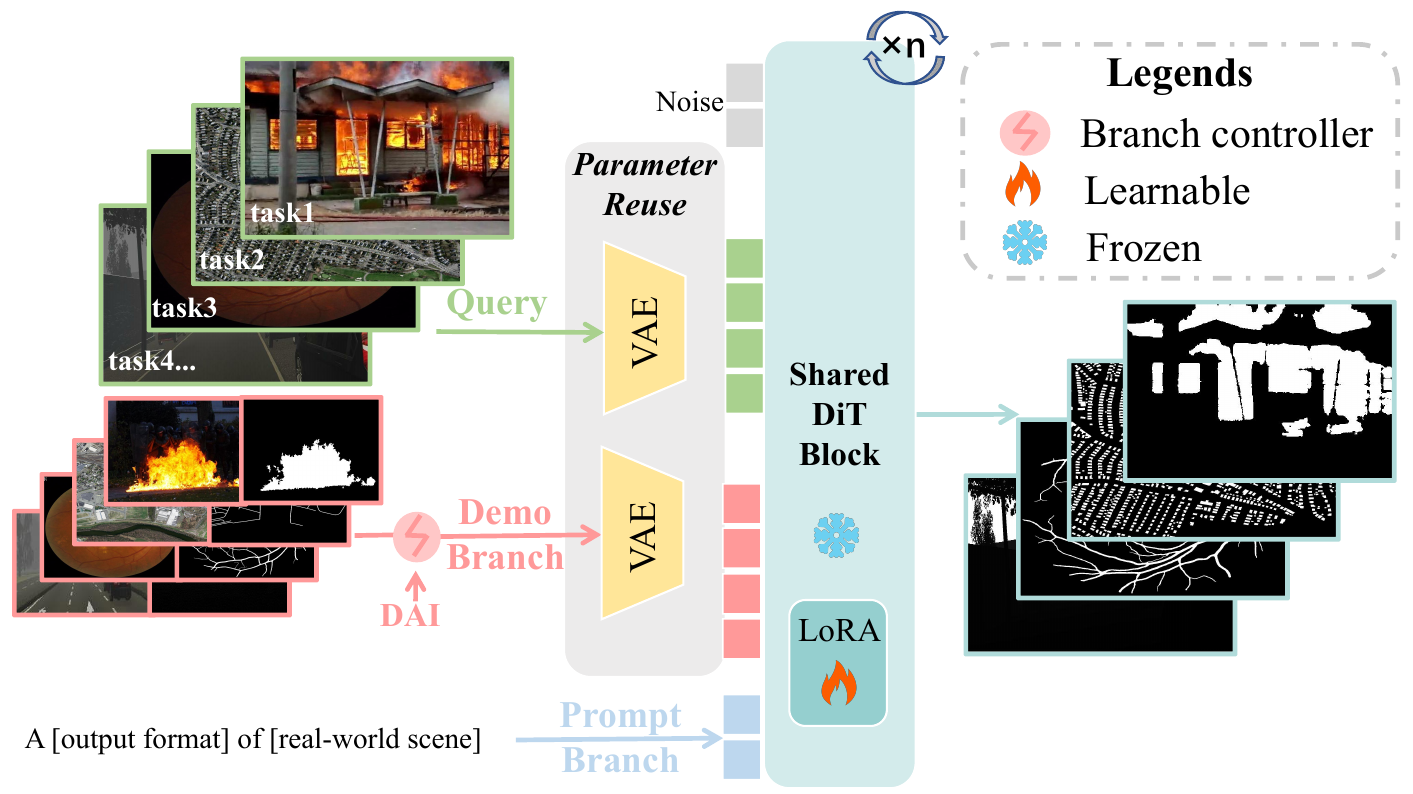}
    \caption{Overview of the DenseDiT architecture. The framework processes a query image through a parameter-reused VAE encoder, while lightweight prompt and demo branches provide contextual cues. These elements interact within the generative backbone, requiring only LoRA-based fine-tuning to achieve dense prediction in complex, data-scarce real-world scenarios.}
  \label{fig:densedit_arch}
\end{figure}

Building upon the DiT architecture, we aim to tackle diverse real-world dense prediction tasks amid inherent data scarcity.
As illustrated in Fig.~\ref{fig:densedit_arch}, we convert the generative backbone into a real-world dense predictor with minimal architectural changes, preserving its pretrained strengths while enhancing its generalization capability across diverse real-world tasks. 
\subsubsection{Parameter-Reuse Mechanism}
Visual priors in DiT are key to DenseDiT’s effectiveness. To preserve them, we perform all the processing in the latent space, where the visual priors are encoded.
Specifically, DenseDiT reuses the VAE to encode the query image into the same latent space as noisy latent, ensuring compatibility with DiT blocks. 
DenseDiT then employs shared DiT blocks for processing. This parameter-reuse mechanism ensures architectural simplicity and low parameter complexity, while maintaining strong compatibility with the pretrained generative backbone. Adaptation to the new input (the query image) is achieved solely through lightweight LoRA~\cite{hu2022lora} modules, adding only 0.1\% parameters.

\subsubsection{Branch-Enhanced Task Understanding}
\label{sec:branch-enhanced}
The tasks in DenseWorld span diverse real-world scenarios, with inputs ranging from natural to non-natural images (e.g., medical images). Although DiT is pretrained on large-scale datasets, adapting its priors to specific tasks under limited supervision remains a challenge. To enhance task understanding with minimal cost, DenseDiT introduces two lightweight branches.

The first is the prompt branch, built directly on DiT’s text encoder. This is a natural design, as DiT is pretrained on image-text pairs, and its text encoder can be reused without further adaptation. For each task, we provide a prompt using the template:
"\textit{A [output format] of [real-world scene]}",
injecting contextual cues for better task alignment.

\begin{table*}[t]
\caption{Comparison with General-Purpose Models: Strong Zero-Shot Baselines for Dense Prediction\label{tab:comparison_general_models}}
\centering
\scriptsize
\setlength{\tabcolsep}{3.5pt}
\fontsize{9pt}{9pt}\selectfont
\begin{tabular}{lccccccc}
\toprule
\textbf{Model} & \textbf{\makecell{Training\\ Samples}} & \textbf{\makecell{Adverse Env.\\ (D-Score)}} & \textbf{\makecell{Smart City.\\ (S-Score)}} & \textbf{\makecell{Medical Assist.\\ (S-Score)}} & \textbf{\makecell{Ecological Mon.\\ (S-Score)}} & \textbf{\makecell{Safety Ctrl.\\ (S-Score)}} & \textbf{Avg.} \\
\midrule
SAM & 1.1B & - & 0.401 & 0.596 & 0.479 & 0.572 & - / 0.512 \\
CLIPSeg & 345K & - & 0.562 & 0.484 & 0.498 & 0.499 & - / 0.511 \\
Grounded-SAM & 1.1B & - & 0.423 & 0.447 & 0.403 & 0.478 & - / 0.438 \\
Marigold & 74K & 0.901 & - & - & - & - & 0.901 / - \\
EcoDepth & 48K & 0.044 & - & - & - & - & 0.044 / - \\
SQLdepth & 24K & 0.352 & - & - & - & - & 0.352 / - \\
ZeoDepth+PF & 19K & 0.577 & - & - & - & - & 0.577 / - \\
Depth-Anything+PF & 19K & 0.771 & - & - & - & - & 0.771 / - \\
\textbf{DenseDiT} & \textbf{15} & \textbf{0.944} & \textbf{0.734} & \textbf{0.825} & \textbf{0.749} & \textbf{0.669} & \textbf{0.944 / 0.744} \\
\rowcolor{gray!10}
w/ Mixed Training & 15$\times$25 & 0.904 & 0.624 & 0.683 & 0.606 & 0.624 & 0.904 / 0.634 \\
\bottomrule
\end{tabular}
\end{table*}


The second is the demonstration branch, designed to mitigate domain gaps for more complex tasks. It employs another parameter-reused VAE to encode a demonstration pair $[I_Q; I_T]$, where $I_Q$ is a query and $I_T$ is its dense label. The resulting latent token interacts with the query latent and prompt token via MMA~\cite{pan2020multi}. To balance performance and computational efficiency, the demonstration branch is selectively activated only when necessary. A binary Distribution Alignment Indicator (DAI) controls demonstration branch activation, indicating whether a task's data distribution aligns with DiT pretraining. This demonstration branch is enabled when DAI = No. DAI labels are automatically assigned using GPT-based classification~\cite{openai2024chatgpt}; specifically, the method evaluates whether a task's imagery aligns with DiT's pretraining distribution or represents a significant domain shift. To ensure robustness, we perform five independent queries and take the majority vote as the final DAI label. The prompt for constructing the DAI label is as follows.

\begin{tcolorbox}[
title={Prompt for obtaining DAI labels},
]
\small
You are given a description and a pair of demonstration images of a dense prediction task. Please determine whether the typical images involved in this task are well-aligned with the training distribution of DiT (e.g., natural, diverse internet images from LAION-5B), or if they represent a significant distribution shift (e.g., medical scans, satellite imagery, infrared).
Please respond with Yes if the task aligns with DiT's training distribution, and No otherwise.
\end{tcolorbox}

This design, combining parameter reuse and branch enhancement, enables DenseDiT to achieve strong multi-task understanding and generalization with only minimal adaptation effort, rather than requiring full training like previous generation-based dense prediction approaches~\cite{ke2024repurposing}.

\subsubsection{Input Representation and Training Objective}
Our training objective is aligned with the DiT framework. Specifically, we optimize the model to predict the velocity field $v_\theta$, minimizing its discrepancy from the true velocity:
\begin{equation}
\label{eq:loss_training}
\text{Loss} = \mathbb{E}_{z_0, t} \left\| v_\theta(z_t, z', t, C_d, C_p) - u(z_t) \right\|_2^2
\end{equation}
where $z_t$ is the noisy latent at timestep $t$, $z'$ is the latent of the query image, and $C_d$, $C_p$ are the latent tokens from the demonstration and prompt branches, respectively.

\section{Experiments}
\subsection{Implementation Details}\label{sec:implementation_details}
DenseDiT is built upon FLUX.1-dev~\cite{flux2024}. By default, we train a separate model for each task. 
Additionally, we train a single model on mixed-task data to further evaluate its multi-task learning capability.
All training is performed using LoRA~\cite{hu2022lora} fine-tuning. We optimize using Prodigy~\cite{mishchenko2023prodigy} with safeguard warmup, bias correction, and a weight decay of 0.01.
Training is conducted on NVIDIA L40S GPUs with a batch size of 1 and gradient accumulation over 8 steps.
\subsection{Main results}
\subsubsection{Comparison with General-Purpose Models}
To evaluate the generalizability of DenseDiT in various real-world scenarios, 
we compare its performance on the DenseWorld benchmark with several strong general-purpose models known for their strong zero-shot capabilities to perform various dense prediction tasks directly.
For pixel-level regression tasks, we select five stable baselines: Marigold~\cite{ke2024repurposing}, EcoDepth~\cite{patni2024ecodepth}, SQLdepth~\cite{wang2024sqldepth}, ZeoDepth~\cite{bhat2023zoedepth}+PF~\cite{li2024patchfusion}, and Depth-Anything~\cite{yang2024depth}+PF. For pixel-level classification tasks, we include three established general-purpose baselines: SAM~\cite{kirillov2023segment}, CLIPSeg~\cite{luddecke2022image}, and Grounded-SAM~\cite{ren2024grounded}.
Table~\ref{tab:comparison_general_models} presents the quantitative results. DenseDiT outperforms all baselines on all tasks in DenseWorld, despite training  under limited supervision. DenseDiT achieves an average D-Score of 0.944 and an average S-Score of 0.744, surpassing the second-best models by 4.8\% and 45.3\%, respectively. 
These results highlight DenseDiT’s strong generalizability across diverse real-world scenarios and its practicality for data-efficient dense prediction given the data scarcity inherent to real-world scenarios.
Detailed metrics for two representative tasks are presented in Table~\ref{tab:Detail Fog Depth Est.} and Table~\ref{tab: Detail Cardiac Screen.} to enhance the interpretability of evaluation results; results for all other tasks are in the supplementary material. These detailed metrics align consistently with the trends reflected in the D/S-Score.

We also train a single DenseDiT model on mixed-task data to assess multi-task learning capability. Although it is slightly weaker than per-task training, a result consistent with the known challenges of multi-task learning across disparate tasks and domains, it still outperforms all baselines. This demonstrates its robustness under limited supervision, a setting that closely reflects real-world application constraints. Certain tasks even benefit from mixed training, indicating inter-task learning potential not achieved in~\cite{zhao2025diception}.

\begin{table}[htbp]
\caption{Evaluation results of detailed metrics for \textit{Fog Depth Est.} task\label{tab:Detail Fog Depth Est.}}
\centering
\resizebox{\linewidth}{!}{
\begin{tabular}{lccccccc}
\toprule
\textbf{Model} & $\delta_1 \uparrow$ & $\delta_2 \uparrow$ & $\delta_3 \uparrow$ & REL $\downarrow$ & Sq-rel $\downarrow$ & RMS $\downarrow$ & RMS log $\downarrow$ \\
\midrule
Marigold                & \underline{0.762} & \underline{0.938} & 0.978 & \underline{0.163} & \underline{1.187} & \underline{6.114} & \underline{0.101} \\
ECoDepth                & 0.481 & 0.776 & 0.915 & 0.368 & 3.579 & 9.657 & 0.160 \\
SQLdepth                 & 0.344 & 0.670 & 0.889 & 0.408 & 3.945 & 10.563 & 0.189 \\
ZoeDepth+PF              & 0.575 & 0.898 & 0.971 & 0.239 & 1.960 & 7.888 & 0.133 \\
Depth-Anything+PF        & 0.588 & 0.932 & \underline{0.984} & 0.223 & 1.702 & 7.804 & 0.110 \\
DenseDiT                 & \textbf{0.845} & \textbf{0.969} & \textbf{0.991} & \textbf{0.139} & \textbf{0.599} & \textbf{3.794} & \textbf{0.077} \\
\rowcolor{gray!10}
w/ Mixed Training & 0.873 & 0.976 & 0.991 & 0.120 & 0.517 & 3.678 & 0.074 \\
\bottomrule
\end{tabular}
}
\end{table}

\begin{table}[htbp]
\caption{Evaluation results of detailed metrics for \textit{Cardiac Screen} task.\label{tab: Detail Cardiac Screen.}}
\centering
\centering
\begin{tabular}{lccc}
\toprule
\textbf{Model} & IoU $\uparrow$ & PA $\uparrow$ & DiCE $\uparrow$ \\
\midrule
SAM & \underline{0.535} & \textbf{0.732} & \textbf{0.652} \\
CLIPSeg & 0.439 & 0.515 & 0.481 \\
Grounded-SAM & 0.402 & 0.478 & 0.449 \\
DenseDiT & \textbf{0.550} & \underline{0.667} & \underline{0.617} \\
\rowcolor{gray!10}
w/ Mixed Training & 0.693 & 0.800 & 0.792 \\
\bottomrule
\end{tabular}
\end{table}

\subsubsection{Comparison with Task-Specific Models}
We observe that two tasks in DenseWorld, \textit{Road Crack Det.} and \textit{Urban Layout Anal.}, have widely studied task-specific models due to their long-standing researches with curated datasets. Such models often perform well within their domains due to architectural customization and large-scale training data, but typically lack generalizability to broader scenarios. To examine this, we compare DenseDiT with state-of-the-art task-specific models on these tasks.
For detailed evaluation, we adopt standard metrics from each domain. As shown in Table~\ref{tab:crack_comparison} and Table~\ref{tab:road_comparison}, DenseDiT outperforms task-specific models in both tasks, despite not relying on task-engineered architecture or large-scale training data. These results highlight the strong performance and generalization of DenseDiT in real-world applications.

\vspace{-0.5em}
\begin{table}[htbp]
\caption{Task-specific comparison on \textit{Road Crack Det.} task\label{tab:crack_comparison}}
\centering
\begin{tabular}{lcccc}
\toprule
\textbf{Model} & IoU$\uparrow$ & PA$\uparrow$ & Dice$\uparrow$ & cIoU$\uparrow$~\cite{benz2024omnicrack30k}\\
\midrule
CrackFormer~\cite{liu2021crackformer}         & 0.512  & 0.729 & 0.058 & 0.198\\
TOPO~\cite{pantoja2022topo}                   & 0.569  & 0.594 & 0.625 & 0.489\\
nnUet~\cite{benz2024omnicrack30k}             & 0.732  & 0.791 & 0.811 & 0.818\\
CT-CrackSeg~\cite{tao2023convolutional}       & 0.677  & 0.770 & 0.758 & 0.603\\
\midrule
DenseDiT & \textbf{0.774} & \textbf{0.863} & \textbf{0.855} & \textbf{0.844} \\
\bottomrule
\end{tabular}
\end{table}


\begin{table}[htbp]
\caption{Task-specific comparison on \textit{Urban Layout Anals.} task\label{tab:road_comparison}}
\centering
\begin{tabular}{lcccc}
\toprule
\textbf{Model} & Recall$\uparrow$ & Precision$\uparrow$ & IoU$\uparrow$ & F1$\uparrow$ \\
\midrule
D-LinkNet~\cite{zhou2018d}   & 0.337 & 0.509 & 0.237 & 0.371 \\
NL-LinkNet~\cite{wang2021nl} & 0.435 & 0.567 & 0.317 & 0.469 \\
RCFSNet~\cite{yang2022road}  & \textbf{0.813} & 0.569 & 0.501 & 0.665 \\
sam-road~\cite{hetang2024segment} & 0.343 & 0.715 & 0.296 & 0.437 \\
\midrule
DenseDiT & 0.619 & \textbf{0.746} & \textbf{0.512} & \textbf{0.672} \\
\bottomrule
\end{tabular}
\end{table}

\subsubsection{Comparison with Fine-tuned Models}
The baselines in Table~\ref{tab:comparison_general_models} are trained on large-scale datasets relevant to their respective tasks (e.g., depth or segmentation). This provides them with substantial prior task understanding and strong zero-shot capabilities to perform various dense prediction tasks directly.
An intuitive yet nontrivial hypothesis is that further fine-tuning these powerful zero-shot models could yield additional performance gains; however, it could also impair their pre-trained task understanding (often referred to as catastrophic forgetting~\cite{kirkpatrick2017overcoming}) or lead to overfitting, thus leading to compromised performance, especially given the extreme data scarcity inherent to real-world applications.
To test this hypothesis and to thoroughly evaluate DenseDiT's performance and generalization, we further compare it against fine-tuned versions of these powerful zero-shot baselines. 
Due to computational constraints, we perform full fine-tuning on the two strongest models, SAM~\cite{kirillov2023segment} and Marigold~\cite{ke2024repurposing}, following their original setups but using 15 training samples per task from DenseWorld.
We also include a recent few-shot model~\cite{hossain2024visual} under the same setting. 
As shown in Table~\ref{tab:fog_comparison} and Table~\ref{tab:crack_comparison_finetune}, the results validate our hypothesis: although all fine-tuned models benefit from additional real-world data, zero-shot baselines fail to achieve robust generalization under extreme data scarcity. In contrast, DenseDiT outperforms them across all metrics. These results demonstrate DenseDiT's exceptional capability in adapting to complex, data-scarce real-world scenarios with minimal data requirements.

\vspace{-0.5em}
\begin{table}[htbp]
\caption{Few-shot comparison on the \textit{Road Crack Det.} task.\label{tab:crack_comparison_finetune}}
\centering
\setlength{\tabcolsep}{4pt}
\fontsize{9pt}{9pt}\selectfont
\small
\begin{tabular}{lccccccc}
\toprule
\textbf{Model} & $\delta_1$↑ & REL↓ & Sq-rel↓ & RMS↓ & RMS log↓ \\
\midrule
Marigold (FT) & 0.796 & 0.145 & 0.974 & 5.712 & 0.089 \\
DenseDiT      &  \textbf{0.845}  & \textbf{0.139} & \textbf{0.599} & \textbf{3.749} & \textbf{0.077} \\
\bottomrule
\end{tabular}
\end{table}

\vspace{-1em}
\begin{table}[htbp]
\caption{Few-shot comparison on the \textit{Fog Depth Est.} task.\label{tab:fog_comparison}}
\centering
\small
\begin{tabular}{lccc}
\toprule
\textbf{Model} & IoU↑ & PA↑ & Dice↑ \\
\midrule
SAM (FT)   & 0.622 & 0.738 &  0.688 \\
Hossain et al. (FT) & 0.625 & 0.655 &  0.689 \\
DenseDiT   & \textbf{0.774}  & \textbf{0.863} & \textbf{0.855} \\
\bottomrule
\end{tabular}
\end{table}

\begin{figure*}[t]
  \centering
  \includegraphics[width=0.95\textwidth]{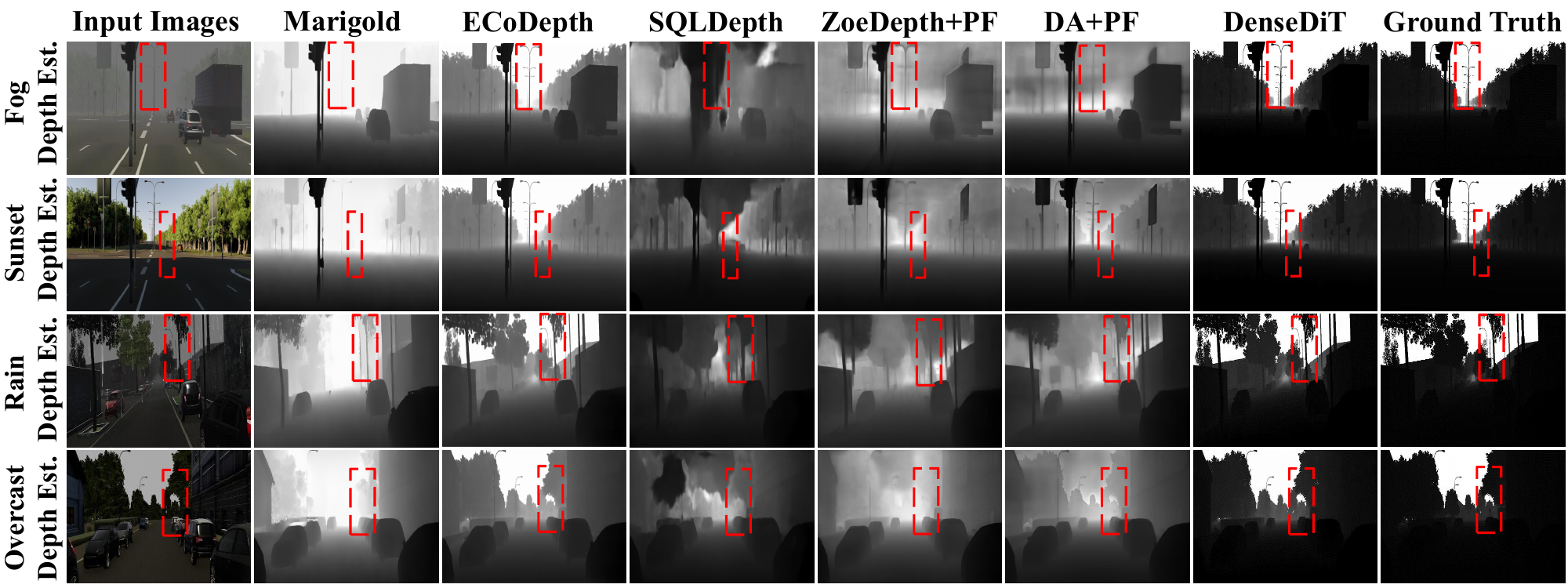}
  \caption{\textbf{Qualitative comparisons on pixel-level regression.} In the first and second row, DenseDiT successfully predicts occluded structures in fog or shadow, highlighting its capability for scene-level reasoning. The third row showcases DenseDiT’s ability to capture fine-grained details such as distant lampposts and layered foliage. The forth row emphasizes its sensitivity to abrupt depth transitions, producing sharper and more consistent boundaries than competing models.}
  \label{fig:depth_visual}
\end{figure*}

\begin{figure}[t]
  \centering
  \includegraphics[width=0.95\linewidth]{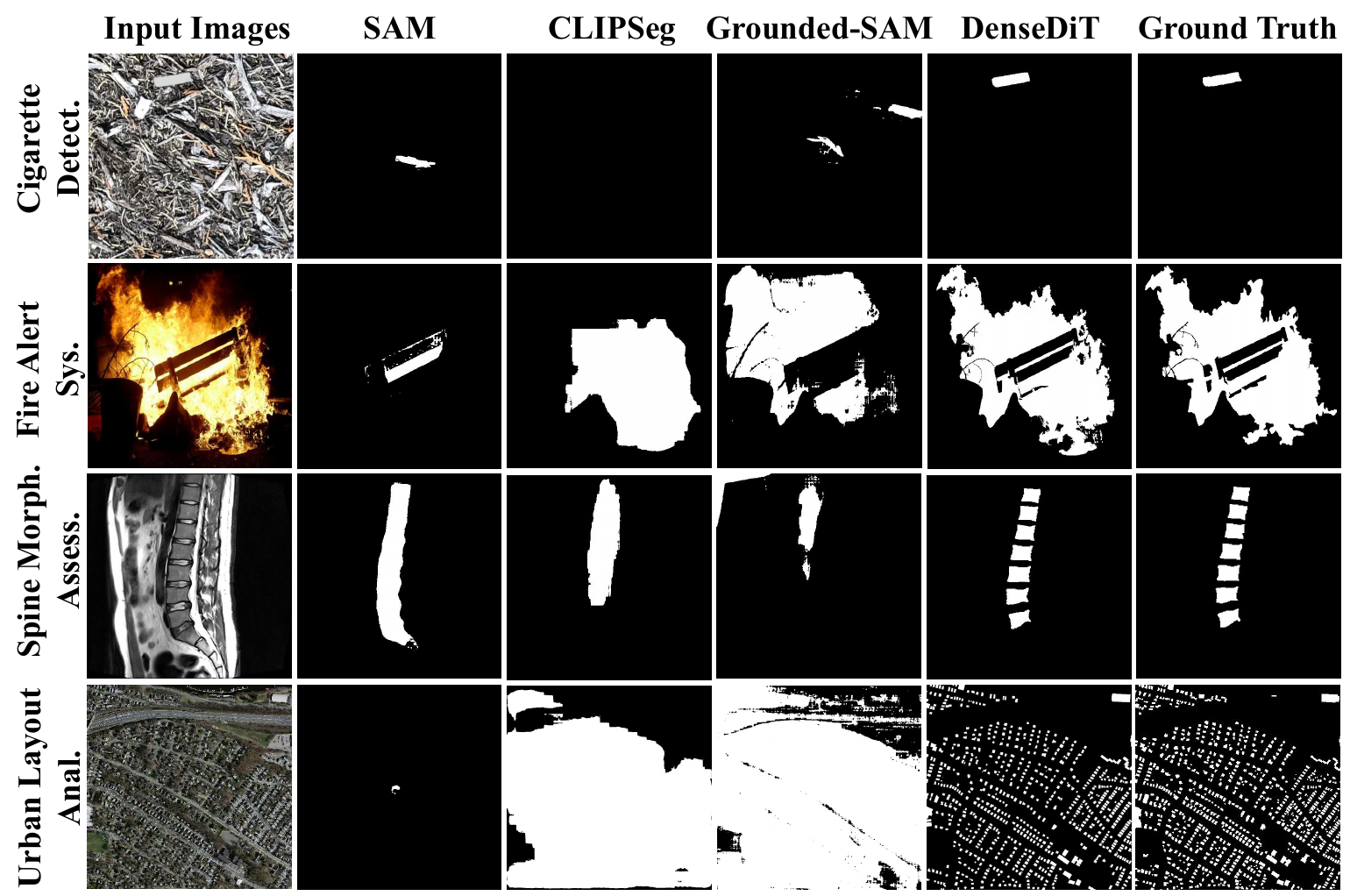}
    \caption{\textbf{Qualitative comparisons on pixel-level classification.} DenseDiT handles cluttered backgrounds (row 1), detects dynamic concepts like fire (row 2), and localizes fine structures in medical/satellite images (rows 3-4).}
  \label{fig:seg_visual}
\end{figure}

\subsubsection{Comparison on Common Benchmarks}
To evaluate DenseDiT beyond DenseWorld, we conduct experiments on three common benchmarks: KITTI~\cite{geiger2013vision}, NYUv2~\cite{silberman2012indoor} and Omnicrack30k~\cite{benz2024omnicrack30k}. DenseDiT is trained on only 15 images from each benchmark’s training set. Table~\ref{tab:kitti_results}, Table~\ref{tab:nyuv2_results} and Table~\ref{tab:omnicrack_results} show that DenseDiT achieves competitive results, compared to the state-of-the-art methods SQLdepth~\cite{wang2024sqldepth} (KITTI),  ECoDepth~\cite{patni2024ecodepth} (NYUv2) and nnUnet~\cite{benz2024omnicrack30k} (Omnicrack30k).

\vspace{-1em}
\begin{table}[htbp]
\caption{Quantitative results on KITTI benchmark.\label{tab:kitti_results}}
\centering
\begin{tabular}{lccccccc}
\toprule
\textbf{Model} & \makecell{Training\\Samples} & $\delta_1 \uparrow$ & REL $\downarrow$ & RMS $\downarrow$ & REL log $\downarrow$ \\
\midrule
ECoDepth & 48k & 0.978 & 0.059 & 0.218 & 0.026 \\
DenseDiT & 15 & 0.928 & 0.084 & 0.317 & 0.035 \\
\bottomrule
\end{tabular}
\end{table}

\begin{table}[htbp]
\caption{Quantitative results on NYUv2 benchmark.\label{tab:nyuv2_results}}
\setlength{\tabcolsep}{2.3pt}
\fontsize{9pt}{9pt}\selectfont
\centering
\begin{tabular}{lccccccc}
\toprule
\textbf{Model} & \makecell{Training\\Samples} & $\delta_1 \uparrow$ & REL $\downarrow$ & Sq-Rel $\downarrow$ & RMS $\downarrow$ & RMS log $\downarrow$ \\
\midrule
SQLdepth & 24k  & 0.928 & 0.082 & 0.607 & 3.914 & 0.160\\
DenseDiT & 15 & 0.851 & 0.129 & 0.589 & 3.918 & 0.072 \\
\bottomrule
\end{tabular}
\end{table}

\begin{table}[htbp]
  \caption{Quantitative results on Omnicrack30k benchmark.\label{tab:omnicrack_results}}
  \centering
  \begin{tabular}{lcccc}
    \toprule
    Method & IoU $\uparrow$ & PA $\uparrow$ & cIoU $\uparrow$ \\
    \midrule
    nnUnet & 0.683 & 0.756 & 0.479 \\
    DenseDiT            & 0.608 & 0.635 & 0.245 \\
    \bottomrule
  \end{tabular}
\end{table}

\subsubsection{Visual Comparisons}
Fig.~\ref{fig:depth_visual} and Fig.~\ref{fig:seg_visual} present qualitative comparisons. As shown in Fig.~\ref{fig:depth_visual}, prior methods often struggle with occlusions or low-visibility (e.g., fog, rain, overcast), 
while DenseDiT yields coherent depth maps by leveraging semantic and visual cues via its prompt and demonstration branches. Fig.~\ref{fig:seg_visual} illustrates DenseDiT's advantage in segmentation tasks. Notably, in the fourth row, DenseDiT preserves structured urban layouts in satellite imagery where other models struggle. While SAM is a strong segmentation baseline, its design prioritizes object-centric cues, leading to suboptimal results in structured or dynamic scenes. More visual comparisons are shown in the supplementary material.

\subsection{Ablation studies}
\subsubsection{Effectiveness of the Demonstration Branch}
As introduced in Section~\ref{sec:branch-enhanced}, DenseDiT employs a demonstration branch to enhance task understanding in complex scenarios, activated when DAI=No.
We validate its effectiveness via ablations on two representative tasks: \textit{Spinal Morphology Assessment}, involving intricate medical structures, and \textit{Instant Fire Alert}, requiring dynamic visual pattern detection. These tasks exemplify real-world challenges where contextual cues are vital.
As shown in Table~\ref{tab:ablation_branch_simplified}, both tasks show notable performance gains with the branch activated, highlighting its impact on generalization in complex settings.

\begin{table}[htbp]
\caption{Effectiveness of the demonstration branch.\label{tab:ablation_branch_simplified}}
\setlength{\tabcolsep}{4pt}
\fontsize{9pt}{9pt}\selectfont
\centering
\begin{tabular}{lcc}
\toprule
\textbf{Method} & \scriptsize \textbf{\makecell{Fire Alert Sys.\\(D/S-Score$\uparrow$)}} & \scriptsize \textbf{\makecell{Spine Morph. Assess.\\(D/S-Score$\uparrow$)}} \\
\midrule
w/o demonstration branch & 0.744 & 0.613 \\
w/ demonstration branch & \textbf{0.876} & \textbf{0.956} \\
\bottomrule
\end{tabular}
\end{table}


\subsubsection{Training Loss}
\begin{figure}[htbp]
\centering
\includegraphics[width=0.8\linewidth,height=0.5\textheight,keepaspectratio]{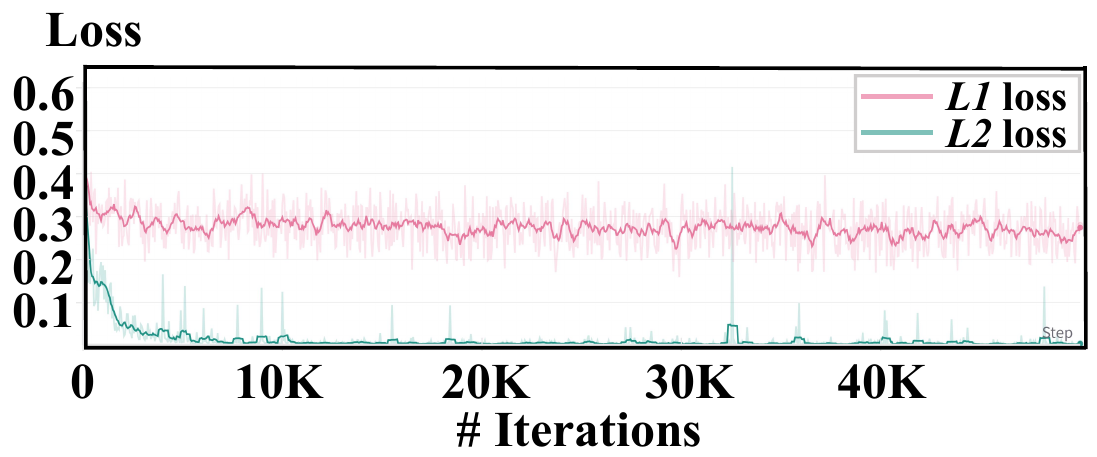}
\caption{Ablation study on loss functions.}
\label{fig:loss}
\end{figure}

DenseDiT is a generative-based framework for real-world dense prediction.
While prior generative-based dense prediction methods~\cite{saxena2023monocular} commonly use \textit{L1} loss for its robustness to noise in early benchmarks~\cite{geiger2013vision}, we find that \textit{L2} loss (Eq.\ref{eq:loss_training}), yields smoother convergence and a more stable plateau in DenseDiT, as shown in Fig.~\ref{fig:loss}.
We attribute this to two factors: (1) the high quality of DenseWorld reduces the need for noise-tolerant objectives, and (2) \textit{L2} loss provides stronger gradient signals for more effective optimization.

\vspace{-1em}

\vspace{-1em}

\subsubsection{The Function of Prompt Branch}
\begin{figure}[htbp]
\centering
\includegraphics[width=0.82\linewidth]{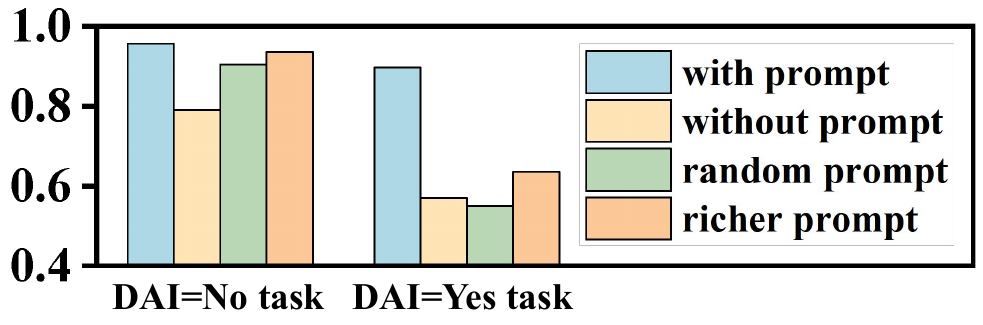}
\caption{Ablation study on prompt branch.}
\label{fig:prompt_branch}
\end{figure}

As introduced in Section~\ref{sec:branch-enhanced}, DenseDiT includes a lightweight prompt branch using the native text encoder to deliver contextual cues via a text template. Since prompt quality is crucial for generative models~\cite{jia2024chatgen}, we evalute its impact in DenseDiT.
Fig.~\ref{fig:prompt_branch} compares four settings: \textit{with prompt}, \textit{without prompt}, \textit{random prompt} (\#\$\%\^{}\&*\!@), and \textit{richer prompt}. We report results on two representative tasks under both DAI=No and DAI=Yes, with similar trends observed across other tasks in DenseWorld.
Prompts consistently improve D/S-Score. In DAI=No tasks, removing or randomizing the prompt degrades performance. The effect is more evident in DAI=Yes tasks, where prompts are the only source of task semantics.
Interestingly, richer prompts reduce performance compared to the text template. We attribute this to DenseDiT’s generative backbone, where overly detailed prompts trigger unnecessary imagination and distract from the core task, reducing prediction accuracy. These results highlight the importance of concise, task-aligned prompts for robust and generalizable dense prediction.

\begin{figure}[t]
\centering
\includegraphics[width=0.8\linewidth,height=0.5\textheight,keepaspectratio]{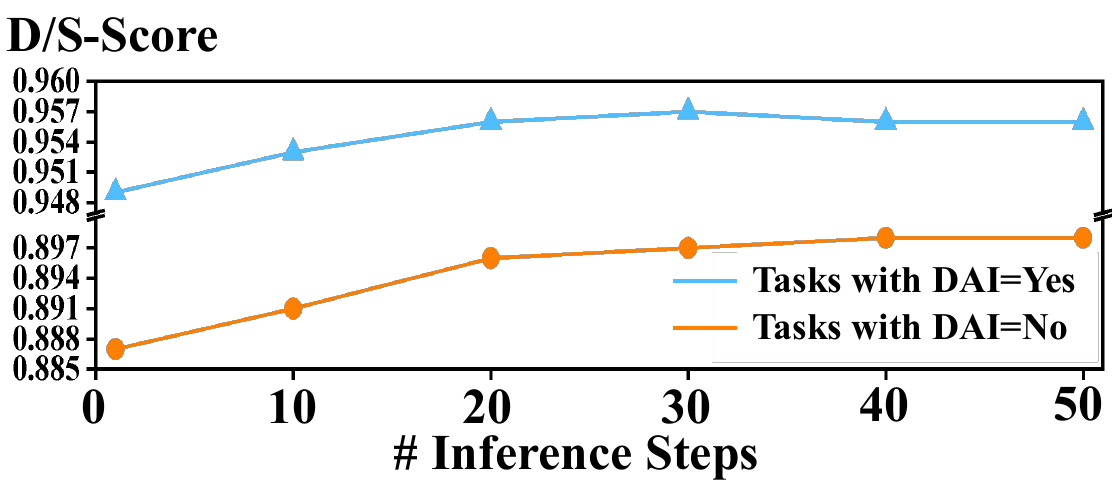}
\caption{Ablation study on inference steps.}
\label{fig:steps}
\vspace{-1em}
\end{figure}

\subsubsection{Inference steps}
Inference steps is key to generative model. More steps improve quality at higher computational cost. As shown in Fig.~\ref{fig:steps}, we evaluate this effect on two tasks under DAI=No and DAI=Yes conditions, with similar trends across other tasks in DenseWorld. DenseDiT performs stably across a wide range of steps, with a noticeable improvement up to 20 steps. This indicates that 20 steps provide a balance between accuracy and efficiency. Notably, for DAI=Yes tasks, performance slightly declines beyond 30 steps, likely due to over-sampling effects degrading fine-grained details~\cite{nichol2021improved}.

\subsection{Generalization Across Backbones}
To validate the generality of our framework and unified training strategy across different backbones, we implement DenseDiT on SD 2.1, which utilizes a convolutional U-Net denoiser, in contrast to the Transformer-based denoiser used in DiT. We report results on two representative tasks: \textit{Automatic Tree Tagging} and \textit{Fog Depth Estimation}; similar trends are observed across other tasks in DenseWorld. As shown in Table~\ref{tab:sd21_fog} and Table~\ref{tab:sd21_tree}, DenseDiT (SD) achieves competitive performance, significantly outperforming all baselines. Although a slight performance gap exists compared to the DiT-based variant, which is expected given the Transformer's superior long-range modeling capacity, these results consistently demonstrate the effectiveness of our framework and unified training strategy in complex, data-scarce real-world scenarios.
\vspace{-1.7em}
\begin{table}[htbp]
\caption{Quantitative results on \textit{Fog Depth Estimation} task.\label{tab:sd21_fog}}
\centering
\resizebox{\linewidth}{!}{
\begin{tabular}{lccccccc}
\toprule
\textbf{Model} & $\delta_1 \uparrow$ & REL $\downarrow$ & Sq-rel $\downarrow$ & RMS $\downarrow$ & RMS log $\downarrow$ \\
\midrule
Marigold & 0.762 & 0.163 & 1.187 & 6.114 & 0.101 \\
ECoDepth & 0.481 & 0.368 & 3.579 & 9.657 & 0.160 \\
SQLdepth & 0.344 & 0.408 & 3.945 & 10.563 & 0.189 \\
ZoeDepth+PF & 0.575 & 0.239 & 1.960 & 7.888 & 0.133 \\
Depth-Anything+PF & 0.588 & 0.223 & 1.702 & 7.804 & 0.110 \\
DenseDiT (DiT) & \textbf{0.845} & \textbf{0.139} & \textbf{0.599} & \textbf{3.794} & \textbf{0.077} \\
DenseDiT (SD) & \underline{0.842} & \underline{0.146} & \underline{0.926} & \underline{4.745} & \underline{0.079} \\
\bottomrule
\end{tabular}
}
\end{table}
\vspace{-1.5em}
\begin{table}[htbp]
\caption{Quantitative results on \textit{Automatic Tree Tagging} task.\label{tab:sd21_tree}}
\centering
\begin{tabular}{lccc}
\toprule
\textbf{Model} & IoU $\uparrow$ & PA $\uparrow$ & DiCE $\uparrow$ \\
\midrule
SAM & 0.765 & 0.840 & 0.809 \\
CLIPSeg & 0.463 & 0.722 & 0.579 \\
Grounded-SAM & 0.391 & 0.529 & 0.457 \\
DenseDiT (DiT) & \textbf{0.858} & \textbf{0.935} & \textbf{0.898} \\
DenseDiT (SD) & \underline{0.845} & \underline{0.916} & \underline{0.885} \\
\bottomrule
\end{tabular}
\end{table}

\vspace{-1em}

\section{Conclusion}
We introduced \textbf{DenseWorld}, a benchmark for unified evaluation of dense prediction across 25 diverse real-world tasks characterized by inherent data scarcity. To address the challenge of learning under such constraints, we proposed \textbf{DenseDiT}, a framework that leverages a parameter-reuse architecture and lightweight branches to achieve robust adaptation with minimal data and additional parameters. Extensive experiments demonstrate that DenseDiT attains strong generalization across complex real-world scenarios, significantly outperforming existing general-purpose and task-specific models. This work provides a significant step towards practical and scalable dense prediction for real-world applications.




\bibliographystyle{IEEEtran}
\bibliography{IEEEabrv,references}

\vfill

\end{document}